\pdfoutput=1

\documentclass[11pt]{article}

\usepackage{EMNLP2022}

\usepackage{times}
\usepackage{latexsym}

\usepackage[T1]{fontenc}

\usepackage[utf8]{inputenc}

\usepackage{microtype}

\usepackage{inconsolata}

\usepackage{amsmath}
\usepackage{bbm}
\usepackage{pifont}
\usepackage{bm}
\usepackage{multirow}
\usepackage{enumitem}
\usepackage{hyperref}
\usepackage{caption}
\usepackage{subcaption}
\usepackage{tablefootnote}
\usepackage{graphicx}
\usepackage{amsfonts} 
\usepackage{textcomp}
\usepackage[linesnumbered,ruled,vlined]{algorithm2e}
\usepackage{multicol,lipsum}
\usepackage{arydshln}
\usepackage[misc]{ifsym}

  {\list{}{\leftmargin=#1\rightmargin=#1}\item[]}%
  {\endlist}

{\noindent\hangindent=6pt\hangafter=1}

\usepackage{tikz}
\usepackage{xcolor}

\newcommand{\emea}{\textsf{EMEA} }

\title{Guiding Neural Entity Alignment with Compatibility}

\author{Bing Liu\textsuperscript{1, \Letter}, Harrisen Scells\textsuperscript{1}, Wen Hua\textsuperscript{1}, Guido Zuccon\textsuperscript{1}, Genghong Zhao\textsuperscript{2}, Xia Zhang\textsuperscript{3} \\
\textsuperscript{1}The University of Queensland, Australia \\
\textsuperscript{2}Neusoft Research of Intelligent Healthcare Technology, Co. Ltd., China\\
\textsuperscript{3}Neusoft Corporation, China\\
\texttt{\{bing.liu, h.scells, w.hua, g.zuccon\}@uq.edu.au} \\
\texttt{\{zhaogenghong,zhangx\}@neusoft.com}}

\begin{document}
\maketitle
\begin{abstract}
Entity Alignment (EA) aims to find equivalent entities between two Knowledge Graphs (KGs). 
While numerous neural EA models have been devised, they are mainly learned using labelled data only.
In this work, we argue that different entities within one KG should have compatible counterparts in the other KG due to the potential dependencies among the entities.
Making compatible predictions thus should be one of the goals of training an EA model along with fitting the labelled data: this aspect however is neglected in current methods. 
To power neural EA models with compatibility, we devise a training framework by addressing three problems: (1) how to measure the compatibility of an EA model; (2) how to inject the property of being compatible into an EA model; (3) how to optimise parameters of the compatibility model.
Extensive experiments on widely-used datasets demonstrate the advantages of integrating compatibility within EA models. In fact, state-of-the-art neural EA models trained within our framework using just 5\% of the labelled data can achieve comparable effectiveness with supervised training using 20\% of the labelled data.  
\end{abstract}

\section{Introduction}
\begin{figure}
    \centering
	\includegraphics[width=8cm]{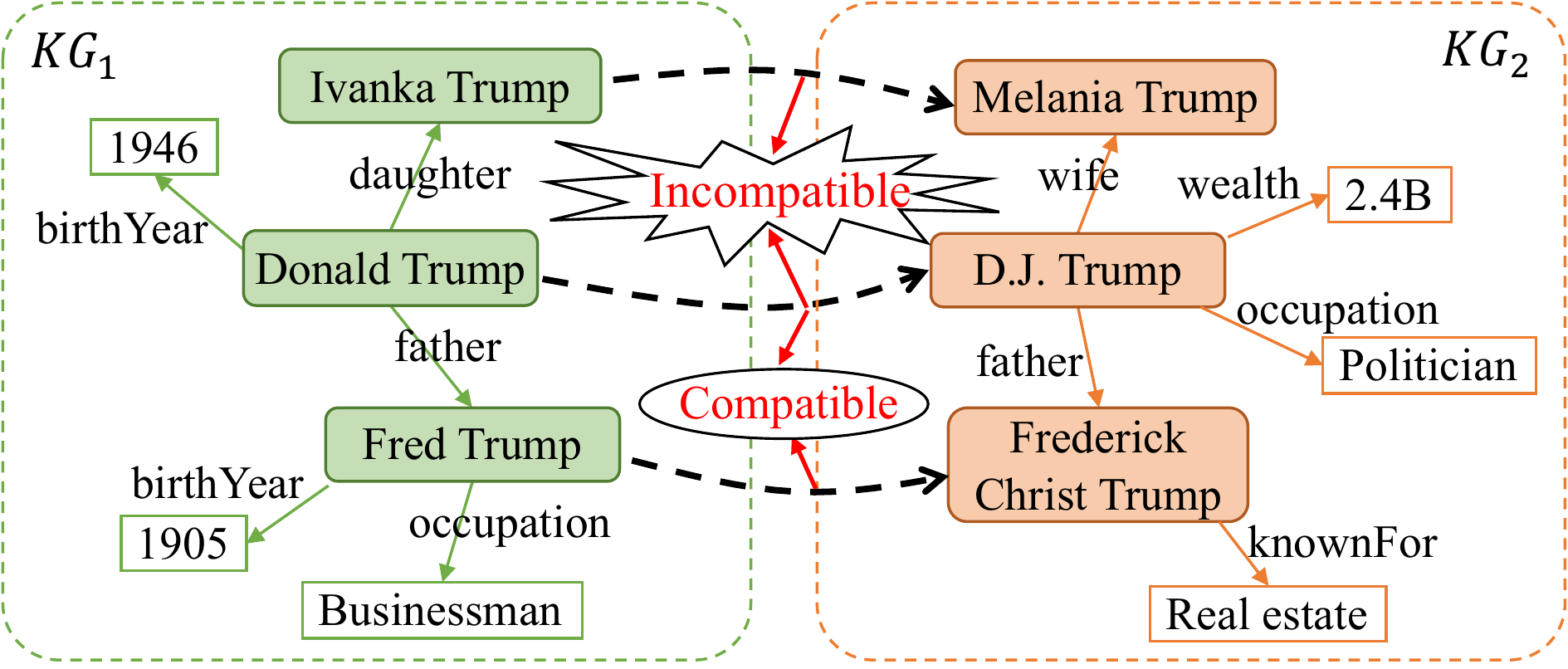}
    \caption{Example of EA predictions. While different mappings have dependencies,  the neural EA model may make incompatible predictions.}
    \label{fig:ea_example}
\end{figure}

Knowledge Graphs (KGs) 
have been widely used 
across many Natural Language Processing applications~\cite{DBLP:journals/tnn/JiPCMY22}. 
However, most KGs suffer from incompleteness which limits their impact on downstream applications. At the same time, different KGs often contain complementary knowledge.
This makes fusing complementary KGs a promising solution for building a more comprehensive KG. 
Entity Alignment (EA), which identifies equivalent entities between two KGs, is essential for KG fusion.
Given the two examples KGs shown in Fig.~\ref{fig:ea_example}, EA aims to recognize two entity mappings $\textit{Donald Trump} \equiv \textit{D.J. Trump}$ and $\textit{Fred Trump} \equiv \textit{Frederick Christ Trump}$.

Neural EA models are the current state-of-the-art for entity alignment~\cite{DBLP:journals/pvldb/SunZHWCAL20,DBLP:journals/tkde/ZhaoZTWS22,DBLP:conf/coling/ZhangLCCLXZ20,DBLP:conf/www/MaoWWL21}
These methods use pre-aligned mappings to learn an EA model: it encodes entities into informative embeddings and then, for each source entity, selects the closest target entity in the vector space as its counterpart. 
Though significant progress has been achieved, the \textit{dependencies} between entities, which is the nature of graph data, is under-explored.
In an EA task, the \textit{counterparts} of different entities within one KG should be \textit{compatible} w.r.t. the underlying dependencies.
For example, in Fig~\ref{fig:ea_example}, the two mappings $\textit{Donald Trump} \equiv \textit{D.J. Trump}$ and $\textit{Ivanka Trump} \equiv \textit{Melania Trump}$ should not co-exist at the same time (i.e. they are incompatible) since \textit{"someone's daughter and wife cannot be the same person".}
On the contrary, $\textit{Donald Trump} \equiv \textit{D.J. Trump}$ and $\textit{Fred Trump} \equiv \textit{Frederick Christ Trump}$ are compatible mappings since equivalent entities' father entities should also be equivalent.
Through an experimental study, we verified that more effective EA models make more compatible predictions (see Appendix~\ref{sec:preliminary} for more details).
Therefore, we argue that making compatible predictions should be one of the objectives of training a neural EA model, other than fitting the labelled data.
Unfortunately, compatibility has thus far been neglected by the existing neural EA works. 

To fill this gap, we propose a training framework \emea, which exploits compatibility to improve existing neural EA models. 
Few critical problems make it challenging to drive a neural EA model with compatibility: 
(1) A first problem is how to measure the overall compatibility of all EA predictions. 
%
%
We notice some reasoning rules defined in traditional reasoning-based EA works~\cite{DBLP:journals/pvldb/SuchanekAS11} can reflect the dependencies between entities well.
To inherit their merits, we devise a compatibility model which can reuse them. 
In this way, we contribute one mechanism of combining reasoning-based and neural EA methods.
%
(2) The second problem is how to improve the compatibility of EA model. 
Compatibility is measured on the counterparts (i.e. labels) sampled from the EA model, but the sampling process is not differentiable and thus the popular approach of regularizing an item in the loss is infeasible. 
We overcome this problem with variational inference.
(3)
The third problem lies in optimising the compatibility model, which has interdependencies with the unknown counterparts.
We solve this problem with a variational EM framework, which alternates updating the neural EA model and the compatibility model until convergence.

Our contributions can be summarized as:
\begin{itemize}[leftmargin=*]
\item We investigate the compatibility issue of the neural EA model, which is critical but so far neglected by the existing neural EA works.
\item We propose one generic framework, which can guide the training of neural EA models with compatibility apart from labelled data.
\item Our framework bridges the gap between neural and reasoning-based EA methods.
\item We empirically show compatibility is very powerful in improving neural EA models, especially when the training data is limited~\footnote{Our code and used data are released at \url{https://github.com/uqbingliu/EMEA}}.
\end{itemize}

\section{Related Work}


\noindent \textbf{Neural EA.}
Entity Alignment is an important task and has been widely studied.
Neural EA~\cite{DBLP:journals/pvldb/SunZHWCAL20,DBLP:journals/tkde/ZhaoZTWS22,DBLP:conf/coling/ZhangLCCLXZ20} is current mainstream direction which emerges with the development of deep learning techniques. 
Various neural architectures have been introduced to encode entities. 
Translation-based KG encoders were explored at the start~\cite{DBLP:conf/ijcai/ChenTYZ17,DBLP:conf/ijcai/ZhuXLS17}. Though these models could capture the structure information, they were not capable of incorporating attribute information.
Graph Convolutional Network (GCN)-based encoders later became the mainstream method because they were flexible in combining different types of information and achieved higher performance~\cite{DBLP:conf/emnlp/WangLLZ18,DBLP:conf/acl/CaoLLLLC19,DBLP:conf/wsdm/MaoWXLW20,DBLP:conf/aaai/SunW0CDZQ20,DBLP:conf/cikm/MaoWXWL20}. 
Neural EA models rely on pre-aligned mappings for training~\cite{DBLP:conf/emnlp/LiuSZHZ21}. To improve EA effectiveness, semi-supervised learning (self-training) was explored to generate pseudo mappings to enrich the training data~\cite{DBLP:conf/ijcai/SunHZQ18,DBLP:conf/www/MaoWWL21}.
Our work aims to complement the existing EA works regardless of their training methods.

Among previous neural EA methods, some were done on KGs with rich attributes and pay attention to exploiting extra information other than  KG structure~\cite{DBLP:conf/ijcai/WuLF0Y019,DBLP:conf/aaai/0001CRC21,DBLP:conf/emnlp/LiuCPLC20,DBLP:conf/emnlp/MaoWWL21,DBLP:conf/ijcai/QiZCCXZZ21}.
Alternatively, some others only focused on designing novel models  to extract better features from the KG structure~\cite{DBLP:conf/ijcai/SunHZQ18,DBLP:conf/aaai/SunW0CDZQ20,DBLP:conf/cikm/MaoWXWL20,DBLP:conf/cikm/LiuHZZZ22} since structure is the most basic information and the proposed method would be more generic. 
We evaluate our method by applying it to models that only consider KG structure, which is a more challenging setting.

\noindent
\textbf{Reasoning-based EA.}
In the reasoning-based EA works~\cite{DBLP:conf/aaai/SaisPR07,DBLP:conf/i3/HoganHD07,DBLP:journals/pvldb/SuchanekAS11}, some rules are defined based on the dependencies between entities.
With the rules, label-level reasoning, i.e. inferring the label of one entity according to other entities' labels instead of its own features, was performed to detect more potential mappings from the pre-aligned ones. 
\textit{Functional} relation (or attribute), which can only have one object for a certain subject, is critical for some reasoning rules~\cite{DBLP:conf/i3/HoganHD07}. 
One rule example is: $\exists r, e_1, e'_1: r(e_1,r,e_2), r(e'_1,r,e'_2), r \textit{ is functional}, e_1 \equiv e'_1 \Rightarrow e_2 \equiv e'_2$.
%
\citeauthor{DBLP:conf/aaai/SaisPR07} proposed to combine multiple properties instead of only using a functional property since the combination of several weak properties can also be functional. 
\citeauthor{hogan2010some} quantified {functional} as {functionality} in a statistical way.
\citeauthor{DBLP:journals/pvldb/SuchanekAS11} inherited the reasoning ideas from previous works and further transformed the logic rules into a probabilistic form in their work named PARIS. 
Apart from \cite{DBLP:journals/pvldb/SuchanekAS11}, few recent works~\cite{DBLP:journals/pvldb/SunZHWCAL20,DBLP:journals/tkde/ZhaoZTWS22} verified PARIS can achieve promising performance.
In this work, we reuse one reasoning rule defined in PARIS because it is  very effective and representative.



\noindent\textbf{Combining Reasoning-based and Neural EA.}
One previous work named \textit{PRASE}~\cite{DBLP:conf/ijcai/QiZCCXZZ21} also explored the combination of neural and reasoning-based EA methods.
It used a neural EA model to measure the similarities between entities and fed these similarities to the reasoning method of PARIS.
Our work provides a different combination mechanism of these two lines of methods.



\section{Notations \& Problem Definition}



Suppose we have two KGs $\mathcal{G}$ and $\mathcal{G}'$ with respective entity sets $E$ and $E'$. 
Each source entity $\mathrm{e} \in E$ corresponds to one counterpart variable $y_{e} \in E'$.
For simplicity, we denote the counterpart variables of a set $E$ of entities as $y_E$ collectively, while use $\hat{y}_{e}$ to represent an assignment of $y_{e}$.
The counterpart variables $y_L$ of labelled entities $L \subset E$ are already known (i.e. $\hat{y}_L$).
EA aims to solve the unknown variables $y_U$ of the unlabelled entities $U \subset E$.

\noindent
One neural EA model measures the similarity $s_\Theta(e, e')$ between each source entity $e \in E$ and each target entity $e' \in E'$, and infers its counterpart via $\hat{y}_e = \arg \max_{e' \in E'} s_\Theta(e, e')$. Here, $\Theta$ represents the parameters of the EA model.

\section{The \emea Framework}

\begin{figure}[!t]
    \centering
    \includegraphics[width=8cm]{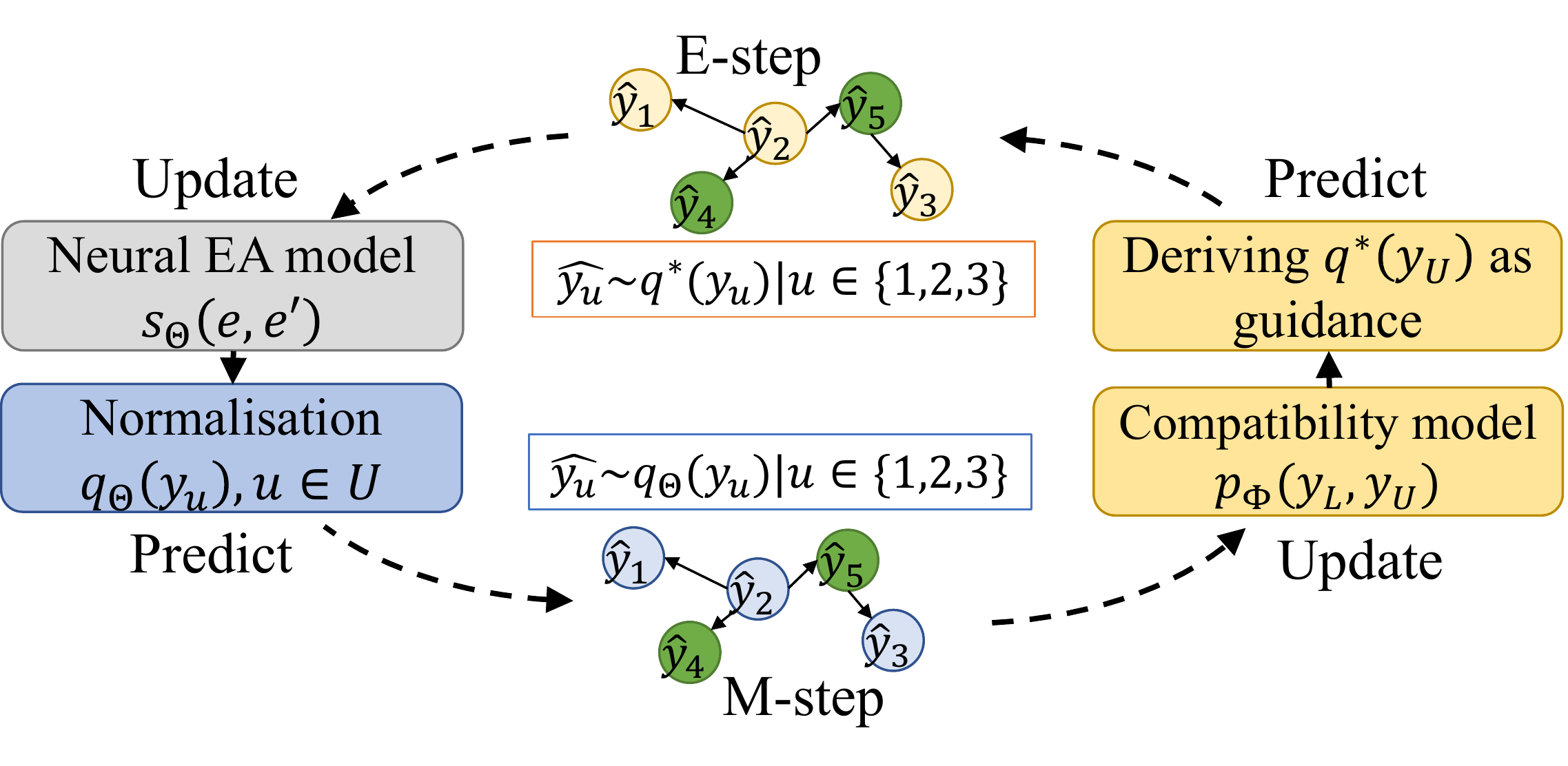}
    \caption{Overview of the \emea framework. The two modules are trained iteratively with variational EM.}
    \label{fig:em_proc}
\end{figure}

Fig.~\ref{fig:em_proc} shows an overview of our \emea framework.
Towards improving a given neural EA model, the \emea performs the following core operations:

\noindent
(1) Normalises similarities between source entity $e$ and all target entities into distribution $q_\Theta (y_e)$;

\noindent
(2) Measure the compatibility of all predictions by modelling the joint probability $p_\Phi(y_L, y_U)$ ($\Phi$ is paramters) of all (known or predicted) mappings;

\noindent
(3) Derive more compatible predictions $q^*(y_U)$ (than current EA model) using the compatibility model to guide updating the EA model.

\noindent
(4)
To learn the parameters of the compatibility model, we devise one optimisation mechanism based on variational EM~\cite{DBLP:books/sp/12/NealH98}, which alternates the update of $\Theta$ and $\Phi$.
The neural EA model $\Theta$ is initially trained in its original way.
In the M-step, we assume current $\Theta$ is correct, and sample $\hat{y}_U \sim q_\Theta$ to update $\Phi$.
In the E-step, we in turn assume current $\Phi$ is correct, and exploit $p_\Phi$ to derive more compatible distribution $q^*(y_u)$. The neural EA model $\Theta$ is then updated using the samples $\hat{y}_U \sim q^*$ together with the labelled data. 
This EM process repeats until $\Theta$ converges. 

\subsection{Normalising EA Similarity to Probability}


Our method relies on the distribution form of counterpart variable $y_e$ as will be seen.
However, the existing neural EA models only output similarities. 
To solve this problem, we introduce a separate model to normalise the similarities into probabilities~\footnote{Some simple normalisation method like MinMax scaler were tried but led to poor results.}. 
Given entity $e \in E$ and similarities $ s_\Theta (e, e')$, we use $ s (e,e') = s_\Theta (e,e')$ and $d (e,e')=\max \left(s_\Theta (e,:)\right)-s_\Theta (e,e')$ as features of each target entity $e' \in E'$. 
These features are combined linearly and fed into a softmax function with a temperature factor $\tau$, as shown in Eq.~\ref{eq:simtoprob_1} and~\ref{eq:simtoprob_2}. The parameters $\Omega = \{ \omega_1,\omega_2,\omega_0, \tau \}$ are learned by minimizing cross-entropy loss on the labelled data, i.e. Eq.~\ref{eq:objective_sim2prob}.
With the obtained model, we can transform $s_\Theta (e,e')$ into $q_\Theta (y_e)$.


\begin{equation}
    f (e,e') = \omega_1 \cdot s (e,e') + \omega_2 \cdot d (e,e') + \omega_0
    \label{eq:simtoprob_1}
\end{equation}

\begin{equation}
    \mathrm{Pr}_{\Omega}(y_e=e') = \frac{\exp(f_\Theta (e,e')/\tau)}{ \mathrm{sum}\left( \exp(f_\Theta (e,:)/\tau) \right)  }
    \label{eq:simtoprob_2}
\end{equation}

\begin{equation}
    O_{\Omega} = - \sum_{e \in L} \log \mathrm{Pr}_{\Omega} (y_e = \hat{y}_e)
    \label{eq:objective_sim2prob}
\end{equation}

\subsection{Measuring Compatibility}\label{sec:self-consistency}

It is not easy to establish the distribution $p_\Phi(y_L, y_U)$ over a large number of variables $y_L, y_U$. To address this problem, we model $p_\Phi(y_L, y_U)$ with graphical model~\cite{DBLP:journals/ftml/WainwrightJ08,bishop2006chapter}, which can be represented by a product of \textit{local functions} (i.e. local compatibility). Each local function only depends on a \textit{factor subset}~\footnote{In graphical model, nodes in $F$ form a factor graph.} $F \subset E$, which is small and can be checked easily.


\subsubsection{Local Compatibility}


One rule $\kappa$ is defined on a set of labels $y_F$ according to the potential dependencies among $y_F$.
The assignments of variables $y_F$ meeting the rule $\kappa$ are thought compatible.
%
Given a rule set $\mathcal{K} = \{ \kappa_1, \kappa_2, ..., \kappa_{|\mathcal{K}|} \}$ and a factor subset $F$ to check, we define the corresponding local compatibility (i.e. local function) as Eq.~\ref{eq:local_func}, where $g_{\kappa}(\cdot)$ is an indicator function, $\Phi=\{ \phi_{\kappa \in \mathcal{K}}, \phi_0 \}$ are the weights of rules.



\begin{equation}
    l (y_F) = \exp \left(\sum_{\kappa \in \mathcal{K}} \phi_\kappa \cdot g_\kappa(y_F) + \phi_0 \right)
    \label{eq:local_func}
\end{equation}

Next, we use two concrete examples to explain local compatibility. 
The PARIS rule is the primary one used by our framework, while another is for exploring the generality of different rule sets.

\paragraph{PARIS Rule}~\cite{DBLP:journals/pvldb/SuchanekAS11}
can be understood intuitively as: \textit{one mapping $y_e = e'$ can be inferred from (or supported by) the other mappings between their neighbours $\mathcal{N}_e$ and $\mathcal{N}_{e'}$}.

\begin{figure}
    \centering
    \includegraphics[width=7.8cm]{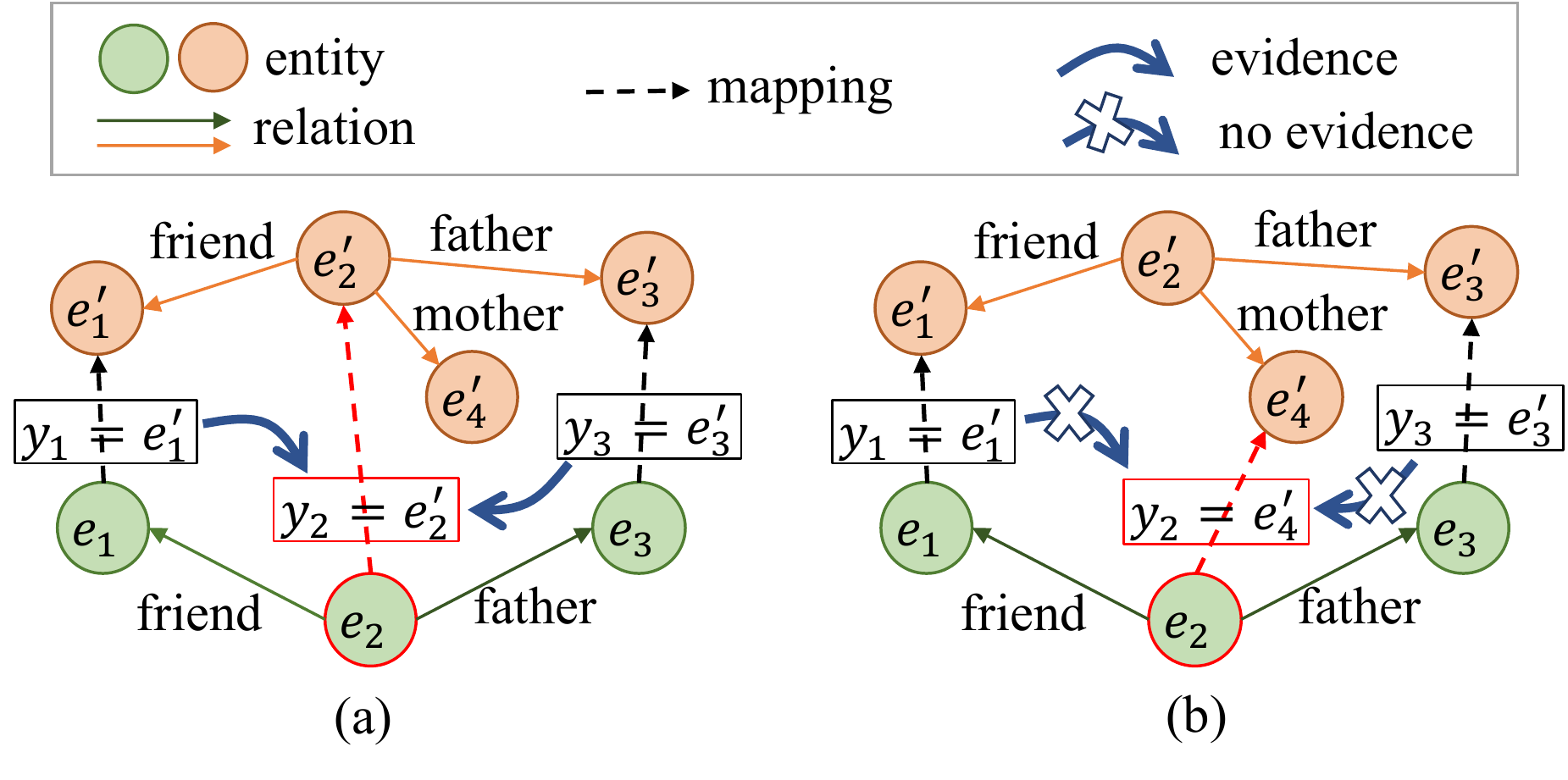}
    \caption{Example of PARIS compatibility. In (a), $y_2=e'_2$ can get supporting evidence from $y_1=e'_1$ and $y_3=e'_3$, while $y_2=e'_4$ in (b) cannot. Thus, the compatibility at $e_2$ in (a) is higher than that in (b).}
    \label{fig:paris_comp}
\end{figure}

Given the predicted mappings, we can build one factor subset $F_e$ at each entity $e$, which contains $e$ and its neighbouring entities $\mathcal{N}_e$. Then, we check whether PARIS rule can be satisfied by mappings $y_{F_e}$.
For example, in Fig.~\ref{fig:paris_comp}, we want to check the PARIS compatibility at $e_2$.
In plot (a), for the mapping $y_2 = e'_2$ we can find two mappings between the neighbours of $e_2$ and $e'_2$ -- $y_1=e'_1$ and $y_3=e'_3$. Also, they can provide supporting evidence for $y_2=e'_2$: if two entities have equivalent father entities and equivalent friend entities, they might also be equivalent.
However, in plot (b), $y_2 = e'_4$ cannot get this kind of supporting evidence since there is no mapping between the neighbours of $e_2$ and $e'_4$.
Thus, the local PARIS compatibility at $e_2$ in plot (a) is higher than that in plot (b).


In this work, we reuse the probabilistic form of PARIS rule (i.e. Eq.(13) in \cite{DBLP:journals/pvldb/SuchanekAS11}) as our indicator function $g$. See Appendix~\ref{app:paris} for its equation with our symbols.

\paragraph{Rules for Avoiding Conflicts.}
We notice that most neural EA works assume that there is no duplicate within one KG, and thus different entities should have different counterparts. Otherwise, EA model makes conflicting predictions for them.

To reduce alignment conflicts, at each entity $e$, we build one factor subset $F_e$, which includes $e$ and its top-$N$ nearest neighbours in the embedding space of EA model.
Basically, $y_e$ should follow the prediction of neural EA model, i.e. $g_1 (y_{F_e}) = \mathbbm{1}_{y_e = \arg \max_{e' \in E} s_\Theta(e, e')}$; Further, $y_e$ should be unique, i.e. $g_2 (y_{F_e}) = \mathbbm{1}_{y_e \neq y_{n}, \forall n \neq e}$.

\subsubsection{Overall Compatibility}


We further formulate the overall compatibility by aggregating local compatibilities on all the factor subsets $\mathcal{F} = \{ F_e, e \in E \}$ as in Eq.~\ref{eq:overall_comp}, where $z$ is for normalisation.

\begin{equation}
    p_\Phi (y_L, y_U) = \frac{1}{z} \prod_{F \in \mathcal{F}} l (y_F)
    \label{eq:overall_comp}
\end{equation}

\begin{equation}
    z = \sum_{y_U \in E^{|U|}} \prod_{F \in \mathcal{F}} l(y_F)
\end{equation}

\noindent
Note that $z$ is intractable because it involves integral over $y_U \in E^{|U|}$, which is a very large space. For such computation reason, we avoid computing $p_\Phi (y_L, y_U)$, conditional probability like $p_\Phi (y_U|y_L)$, and marginal probability $p_\Phi (y_L)$ directly in the following sections. 

Instead, we will exploit $p_\Phi(y_e|y_{-e})$ ($-e$ refers to $E \setminus e$), whose computation is actually much easier.
As in Eq.~\ref{eq:cond_p}, computing $p_\Phi(y_e|y_{-e})$ only involves a few factor subsets containing $e$. 
$\mathrm{MB}^e$ is the Markov Blanket of $e$, which only contains entities cooccuring in any factor subset with $e$.
See Appendix~\ref{app:p_yi} for the derivation process.



\begin{equation}
    \begin{aligned}
        p_\Phi(y_e | y_{-e})
        & = \frac{\prod_{F | e \in F} l(y_F) }{\sum_{e' \in E'} \prod_{F | e \in F} l(y_F | y_e=e') } \\
        & \doteq p_\Phi(y_e | y_{\mathrm{MB}^e})
    \end{aligned}
    \label{eq:cond_p}
\end{equation}

\subsection{Guiding Neural EA with Compatibility}

Suppose we have $p_\Phi (y_L, y_U)$ which can measure the compatibility well. 
We attempt to make the distribution $q_\Theta (y_U)$ close to $p_\Phi(y_U | y_L)$, so that variable $y_U$ sampled from $q_\Theta$ are compatible.
To this end, we treat minimizing the KL-divergence $\mathrm{KL}(q_\Theta(y_U) || p_\Phi(y_U|y_L))$ as one of the objectives of optimising neural EA model $\Theta$.

However, it is difficult to minimize the KL-divergence directly. We solve this problem with variational inference~\cite{DBLP:journals/ijon/Hogan02}.
As shown in Eq.~\ref{eq:KL} and \ref{eq:elbo}, the KL-divergence can be written as the difference between observed evidence $\log p_\Phi (y_L)$, which is irrelevant to $\Theta$, and its Evidence Lower Bound (ELBO), i.e. Eq.~\ref{eq:elbo}
(see Appendix~\ref{app:elbo} for derivation process.).
Minimizing the KL divergence is equivalent to maximizing the ELBO, which is computationally simpler.

\begin{equation}
    \begin{aligned}
        \mathrm{KL}(q_\Theta(y_U) || p_\Phi(y_U|y_L)) =  \log p_\Phi(y_L) 
        - \mathrm{ELBO}
    \end{aligned}
    \label{eq:KL}
\end{equation}

\begin{equation}
    \begin{split}
        \mathrm{ELBO} = & \mathbb{E}_{q_\Theta (y_U)} \log p_\Phi(y_U, y_L) \\ & - \mathbb{E}_{q_\Theta (y_U)} \log q_\Theta (y_U)
    \end{split}
    \label{eq:elbo}
\end{equation}

\noindent
Because $y_i$ are independent in neural EA model, we have $q(y_U) = \prod_{i \in U} q(y_i) $.
In addition, we use pseudolikelihood~\cite{besag1975statistical} to approximate the joint probability $p_\Phi(y_U, y_L)$ for simpler computation, as in Eq.~\ref{eq:pseudolikelihood}. 
Then, ELBO can be approximated with Eq.~\ref{eq:elbo_approx} (see Appendix~\ref{app:Q_proc} for derivation details), where $-u$ denotes $U \setminus u$.

\begin{equation}
    \begin{aligned}
        p_\Phi(y_U, y_L) & = p_\Phi(y_L) \prod_{u \in U} p_\Phi(y_{u} | y_{1:u-1}, y_L )  \\
& \approx p_\Phi(y_L) \prod_{u \in U} p_\Phi(y_u | y_{-u} ) 
    \end{aligned}
\label{eq:pseudolikelihood}
\end{equation}

\begin{equation}
    \begin{split}
        O_\Theta = \sum_{u \in U} \mathbb{E}_{q_\Theta(y_u)} 
         \Big[ \mathbb{E}_{q_\Theta({y_{-u }})}  
         [ \log p_\Phi(y_u | y_{-u }) ]    \\
         - \log q_\Theta(y_u)   \Big]
    \end{split}
    \label{eq:elbo_approx}
\end{equation}

Now, our goal becomes to maximize $O_\Theta$ w.r.t. $\Theta$. 
Our solution is to derive a local optima $q^* (y_U)$ of $q_\Theta (y_U)$ with coordinate ascent, and then exploit $\hat{y}_U \sim q^* (y_U)$ to update $\Theta$. 
In particular, we initialize $q^*(y_U)$ with current $q_\Theta (y_U)$ firstly.
Then, we update $q^*(y_u)$ for each $u \in E$ in turn iteratively. Everytime we only update a single (or a block of) $q^*(y_u)$ with Eq.~\ref{eq:q_star}, which can be derived from $\frac{d Q_\Theta}{d q_\Theta(y_u)} = 0$ (see Appendix~\ref{app:q_star_proc} for derivation details), while keeping the other $q^*(y_{-u})$ fixed. 
This process ends until $q^*(y_U)$ converges.

\begin{equation}
    q^*(y_u) \propto \exp \left(\mathbb{E}_{q_\Theta(y_{\mathrm{MB}^u })}  \log p_\Phi(y_u | y_{\mathrm{MB}^u })  \right)
    \label{eq:q_star}
\end{equation}

\noindent
Afterwards, we sample $\hat{y}_u \sim q^* (y_u)$ for $u \in U$, and join $\hat{y}_U$ with the labelled data $\hat{y}_L$ to form the training data. Eventually, we update $\Theta$ with the original training method of neural EA model.

\subsection{Optimisation with Variational EM}

Though we have derived a way of guiding the training of $\Theta$ with $p_\Phi$, it remains a problem to optimise the weights of rules $\Phi$.
Typically, we learn $\Phi$ by maximizing the log-likelihood of observed data $\log p_\Phi (y_L)$. 
However, as shown in $\log p_\Phi(y_L) = \log \sum_{y_U} p_\Phi(y_L, y_U) $, $\log p_\Phi(y_L)$ relies on the latent variables $y_U$.
We apply a variational EM framework~\cite{DBLP:books/sp/12/NealH98,DBLP:conf/nips/Qu019} to update $\Theta$ and $\Phi$ by turns iteratively.
In E-step, we compute the expectation of $\log p_\Phi(y_L, y_{U})$, i.e. $\mathbb{E}_{p_\Phi(y_U|y_L)} \log p_\Phi(y_L, y_U)$.
Here, we approximate $p_\Phi (y_U|y_L)$ with $q_\Theta (y_U)$ and use the pseudolikelihood to approximate $p_\Phi(y_L, y_U)$; 
Accordingly, we obtain the objective Eq.~\ref{eq:objective_phi} for optimization.
In M-step, we update $\Phi$ to maximize $O_{\Phi}$.

\begin{equation}
    \begin{aligned}
        O_{\Phi} & = \mathbb{E}_{q_\Theta(y_U)}  \log p_\Phi(y_L, y_U)  \\
    & \approx \mathbb{E}_{q_\Theta(y_U)}  \sum_{e \in E} \log p_\Phi( y_e | y_{-e})  
    \end{aligned}
    \label{eq:objective_phi} 
\end{equation}

\subsection{Implementation}

\begin{algorithm}[t!]
    \caption{The \emea Framework}
    \label{alg}
    Train neural EA model $\Theta$ using $\hat{y}_L$ \;
    Normalise EA similarity to get $q_\Theta (y_u)$ \; 
    \For{iterations}{ 
        \tcp{M-step}
        Update $\Phi$ by maximizing Eq.~\ref{eq:objective_phi} \;
        \tcp{E-step}
        Derive $q^*(y_U)$ with Eq.~\ref{eq:q_star} \;
        Sample $\hat{y}_{u} \sim q^*(y_u)$ for $u\in U$ \;
        Update EA model $\Theta$ with data $\hat{y}_U \cup \hat{y}_L$ \;
        Normalise EA similarity to get $q_\Theta (y_u)$  \;
    }
\end{algorithm}

We take a few measures to simplify the computation.
(1) In Eq.~\ref{eq:q_star} and Eq.~\ref{eq:objective_phi}, it is costly to estimate distribution $q^*(y_u)$ and $p_\Phi(y_u)$ because $y_u$'s  assignment space $E'$ can be very large. 
Instead, we only estimate $q^*(y_u)$ for the top $K$ most likely candidates according to current $q_\Theta (y_u)$.
(2) Both Eq.~\ref{eq:elbo_approx} and Eq.~\ref{eq:q_star} involve sampling from $q_\Theta(y_u)$ for estimating the expectation. We only sample one $y_u$ as in $\hat{y}_u = \arg \max_{e' \in E} q_\Theta(y_u=e')$ for each $u \in U$.
(3) When computing $q^*(y_U)$ with coordinate ascent, we treat $U$ as a single block and update $q^*(y_U)$ for once.

\noindent
We describe the whole process of \emea in Alg.~\ref{alg}.

\section{Experimental Settings}


\subsection{Datasets and Partitions}

We choose five datasets widely used in previous EA research. Each dataset contains two KGs and a set of pre-aligned entity mappings.
Three datasets are from \textit{DBP15K}~\cite{DBLP:conf/semweb/SunHL17}, which contains three cross-lingual KGs extracted from DBpedia: French-English (\textit{fr\_en}), Chinese-English (\textit{zh\_en}), and Japanese-English (\textit{ja\_en}). 
Each KG contains around 20K entities, among which 15K are pre-aligned.
The other two datasets are from \textit{DWY100K}~\cite{DBLP:conf/ijcai/SunHZQ18}, which consists of two mono-lingual datasets: \textit{dbp\_yg} extracted from DBpedia and Yago, and \textit{dbp\_wd} extracted from DBpedia and Wikidata. 
Each KG contains 100K entities which are all pre-aligned.
Our experiment settings only consider the structural information of KGs and thus will not be affected by the problems of attributes like name bias in these datasets~\cite{DBLP:journals/tkde/ZhaoZTWS22,DBLP:conf/emnlp/LiuCPLC20}.

Most existing EA works use 30\% of the pre-aligned mappings as training data, which however was pointed out unrealistic in practice~\cite{DBLP:conf/coling/ZhangLCCLXZ20}. 
We explore the power of compatibility under different amounts of labelled data -- 1\%, 5\%, 10\%, 20\%, and 30\% of pre-aligned mappings, which are sampled randomly. 
Another 100 mappings are used as the validation set, while all the remaining mappings form the test set.

\begin{table*}
    \centering
    \footnotesize
    \caption{Overall performance of \textsf{EMEA} and PRASE in combining RREA (sup) and PARIS rule across different percentages (1\%-30\%) of annotations. Bold indicates best for the specific annotation percentage; all differences between RREA and other baselines are statistically significant ($p<0.01$); the hyphen '-' means not applicable because the corresponding methods do not formulate EA as a ranking problem. The results of PRASE and PARIS have big differences from those in the literature because of different experimental settings.}
    \scalebox{0.77}[0.77]{
    
\begin{tabular}{|c|c|c|c|c|c|c|c|c|c|c|c|c|c|c|c|c|}
    \hline
    \multirow{2}{*}{} & \multirow{2}{*}{Method} &\multicolumn{3}{c|}{zh\_en}&\multicolumn{3}{c|}{fr\_en}&\multicolumn{3}{c|}{ja\_en}&\multicolumn{3}{c|}{dbp\_wd}&\multicolumn{3}{c|}{dbp\_yg}\\
     & & Hit@1&MRR&MR&Hit@1&MRR&MR&Hit@1&MRR&MR&Hit@1&MRR&MR&Hit@1&MRR&MR\\
    \hline
    \multirow{4}{*}{1\%} & PARIS & 0.01 &-&-& 0.016 &-&-& 0.002 &-&-& 0.19	 &-&-& 0.451 &-&- \\
    & RREA (sup) & 0.140&0.215&652.3&0.126&0.208&366.5&0.138&0.203&684.0&0.278&0.368&317.9&0.509&0.602&64.4\\
    \cdashline{2-17}
    & PRASE & 0.241	&-&-& 0.227 &-&-&0.163 &-&-& 0.517 &-&-& 0.667 &-&- \\
     & EMEA & \textbf{0.517}&\textbf{0.591}&\textbf{116.4}&\textbf{0.480}&\textbf{0.565}&\textbf{72.1}&\textbf{0.411}&\textbf{0.488}&\textbf{181.3}&\textbf{0.581}&\textbf{0.657}&\textbf{72.8}&\textbf{0.773}&\textbf{0.828}&\textbf{17.6}\\
    \hline
    \multirow{4}{*}{5\%} & PARIS & 0.221&-&-&0.281&-&-&0.226&-&-&0.537&-&-&0.608 &-&- \\
    & RREA (sup)& 0.413&0.518&118.8&0.424&0.539&65.3&0.391&0.496&113.9&0.522&0.616&78.4&0.737&0.803&21.0\\
    \cdashline{2-17}
    & PRASE & 0.461	&-&-& 0.514	&-&-& 0.432 &-&-& 0.531 &-&-& 0.689 &-&- \\
     & EMEA & \textbf{0.665}&\textbf{0.738}&\textbf{36.8}&\textbf{0.677}&\textbf{0.757}&\textbf{18.0}&\textbf{0.630}&\textbf{0.710}&\textbf{35.9}&\textbf{0.708}&\textbf{0.778}&\textbf{21.5}&\textbf{0.811}&\textbf{0.861}&\textbf{12.7}\\
    \hline
    \multirow{4}{*}{10\%} & PARIS & 0.414&-&-&0.473&-&-&0.395&-&-&0.623&-&-&0.64&-&-\\
    & RREA (sup)& 0.542&0.641&56.9&0.571&0.675&31.3&0.528&0.631&52.6&0.622&0.709&36.2&0.782&0.841&14.0\\
    \cdashline{2-17}
    & PRASE & 0.522	&-&-&0.575	&-&-&0.508&-&-& 0.679 &-&-& 0.701 &-&-\\
     & EMEA & \textbf{0.706}&\textbf{0.777}&\textbf{27.4}&\textbf{0.727}&\textbf{0.802}&\textbf{9.7}&\textbf{0.688}&\textbf{0.764}&\textbf{24.5}&\textbf{0.755}&\textbf{0.820}&\textbf{13.4}&\textbf{0.828}&\textbf{0.877}&\textbf{9.2}\\
    \hline
    \multirow{4}{*}{20\%} & PARIS & 0.532 &-&-& 0.584 &-&-& 0.511 &-&-& 0.69 &-&-& 0.676 &-&- \\
    & RREA (sup)& 0.657&0.745&26.5&0.686&0.775&14.7&0.649&0.740&25.3&0.711&0.787&20.6&0.824&0.875&12.0\\
    \cdashline{2-17}
    & PRASE & 0.593 &-&-& 0.622 &-&-& 0.580 &-&-& 0.726 &-&-& 0.719 &-&- \\
     & EMEA & \textbf{0.748}&\textbf{0.815}&\textbf{16.6}&\textbf{0.773}&\textbf{0.841}&\textbf{6.8}&\textbf{0.736}&\textbf{0.807}&\textbf{16.3}&\textbf{0.808}&\textbf{0.866}&\textbf{6.6}&\textbf{0.846}&\textbf{0.891}&\textbf{10.4}\\
    \hline
    \multirow{4}{*}{30\%} & PARIS &  0.589 &-&-& 0.628 &-&-& 0.577 &-&-& 0.739 &-&-& 0.696 &-&- \\
    & RREA (sup)& 0.720&0.797&16.7&0.742&0.821&9.2&0.717&0.797&14.9&0.758&0.827&14.0&0.849&0.894&6.9\\
    \cdashline{2-17}
    & PRASE & 0.623 &-&-& 0.649 &-&-& 0.613 &-&-& 0.754 &-&-& 0.735 &-&- \\
     & EMEA & \textbf{0.782}&\textbf{0.842}&\textbf{12.6}&\textbf{0.801}&\textbf{0.863}&\textbf{5.9}&\textbf{0.771}&\textbf{0.837}&$\textbf{12.6}$&\textbf{0.836}&\textbf{0.889}&\textbf{7.3}&\textbf{0.862}&\textbf{0.904}&\textbf{5.8}\\
    \hline

\end{tabular}

    }
    \label{tab:overall_perf}
\end{table*}

\subsection{Metrics}

EA methods typically output a ranked list of candidate counterparts for each entity. Therefore, we choose metrics for measuring the quality of ranking.
We use \textit{Hit@1} (i.e., accuracy), \textit{Mean Reciprocal Rank (MRR)} and \textit{Mean Rank (MR)} to reflect the model performance at suggesting a single entity, a handful of entities, and many entities. 
%
%
Higher Hit@1, higher MRR, and lower MR indicate better performance. 
Statistical significance is performed using paired two-tailed t-test.

\subsection{Comparable Methods}

\noindent
\textbf{Baselines.}
We select baselines with the following considerations:
(1) To examine the effect of compatibility, we compare \emea with the original neural EA model.
%
(2) We compare \emea with PARIS, which performs reasoning with the rule, to gain insights on different ways of using the rules.
(3) To compare different combination mechanisms of neural and reasoning-based EA methods, we add PRASE~\cite{DBLP:conf/ijcai/QiZCCXZZ21}, which exploits neural models to improve PARIS~\cite{DBLP:journals/pvldb/SuchanekAS11}, as one baseline. 

\noindent
\textbf{Neural EA models.}
For our choice of neural models, we select RREA~\cite{DBLP:conf/cikm/MaoWXWL20}, which is a SOTA neural EA model under both supervised (denoted as RREA (sup)) and semi-supervised (denoted as RREA (semi)) modes.
In addition, we also choose Dual-AMN~\cite{DBLP:conf/www/MaoWWL21}, AliNet~\cite{DBLP:conf/aaai/SunW0CDZQ20} and IPTransE~\cite{DBLP:conf/ijcai/ZhuXLS17} to verify the generality of \emea across different neural models. These three neural models vary in performance (see Appendix~\ref{app:neural_ea_models}) and KG encoders.

Note direct comparison between \emea and the existing neural EA methods is not fair.
The \emea is a training framework designed to enhance the existing neural EA models.
Its effectiveness is reflected by the performance difference of neural EA models before and after being enhanced with \emea. 
The details about reproducibility (e.g. hyperparameter settings, etc.) can be found in Appendix~\ref{app:exp}.


\section{Results}


\paragraph{Comparison with Baselines}
In Table~\ref{tab:overall_perf}, we report the overall performance of \emea with supervised RREA and the baselines. 
Note the results of PRASE and PARIS are much lower than those in the literature because we only use the structure information of KGs for all the methods.
We have the following findings:

(1) By comparing \emea with RREA, we can see that \emea can significantly improve RREA across all the datasets and percentages of labelled data, especially when the amount of labelled data is small. For instance, \emea using 5\% of labelled data can achieve comparable effectiveness with supervised RREA using 20\% of labelled data.


(2) \emea always outperforms PARIS with a big margin. Thus, \emea provides a better way of using the same reasoning rule. PARIS can only do label-level inference based on the reasoning rule, while \emea can combine the power of neural EA model and reasoning rule.

(3) Some existing works show PARIS have very competitive performance with the SOTA neural EA models when the attribute information can be used. However, we find its performance is actually much worse than the SOTA neural model RREA when only the KG structure is available. This is a complementary finding about PARIS to the literature.

(4) Regarding the combination of RREA and PARIS, \emea outperforms PRASE across all datasets and annotation costs.
Though PRASE can always improve PARIS, there are some cases where it is worse than only using RREA. The potential reason is PARIS becomes the bottleneck of PRASE.
On the contrary, \emea is more robust -- it consistently performs better than separately running RREA or PARIS. 

To conclude, \emea can significantly improve the SOTA EA model RREA by introducing compatibility. It also provides an effective combination mechanism of neural and reasoning-based EA methods, which outperforms the existing methods.

The further explorations of \emea are done on \textit{zh\_en} dataset if not specially clarified.

\begin{table}
    \caption{Overall performance of \textsf{EMEA} in combining RREA (semi) with PARIS rule across different percentages (1\%-30\%) of annotations. Bold indicates best for the specific annotation percentage; all differences between RREA and \textsf{EMEA} are statistically significant ($p<0.05$).}
    \scalebox{0.66}[0.66]{
    \begin{tabular}{|c|c|c|c|c|c|c|c|}
    \hline
    \multirow{2}{*}{} & \multirow{2}{*}{Method} &\multicolumn{2}{c|}{zh\_en}&\multicolumn{2}{c|}{fr\_en}&\multicolumn{2}{c|}{ja\_en}\\
     & & Hit@1&MRR&Hit@1&MRR&Hit@1&MRR\\
    \hline
    \multirow{2}{*}{1\%} & RREA (semi) & 0.309&0.405&0.277&0.382&0.263&0.348\\
     & EMEA & \textbf{0.471}&\textbf{0.541}&\textbf{0.435}&\textbf{0.518}&\textbf{0.386}&\textbf{0.457}\\
    \hline
    \multirow{2}{*}{5\%} & RREA (semi)& 0.590&0.683&0.610&0.708&0.550&0.647\\
     & EMEA & \textbf{0.680}&\textbf{0.751}&\textbf{0.703}&\textbf{0.778}&\textbf{0.641}&\textbf{0.716}\\
    \hline
    \multirow{2}{*}{10\%} & RREA (semi)& 0.677&0.757&0.710&0.791&0.658&0.743\\
     & EMEA & \textbf{0.731}&\textbf{0.796}&\textbf{0.762}&\textbf{0.828}&\textbf{0.711}&\textbf{0.781}\\
    \hline
    \multirow{2}{*}{20\%} & RREA (semi)& 0.756&0.821&0.782&0.848&0.743&0.814\\
     & EMEA & \textbf{0.782}&\textbf{0.840}&\textbf{0.805}&\textbf{0.865}&\textbf{0.765}&\textbf{0.829}\\
    \hline
    \multirow{2}{*}{30\%} & RREA (semi)& 0.794&0.851&0.819&0.876&0.790&0.851\\
     & EMEA & \textbf{0.808}&\textbf{0.862}&\textbf{0.833}&\textbf{0.886}&\textbf{0.801}&\textbf{0.859}\\
    \hline
\end{tabular}

    }
    \label{tab:overall_perf_iterative}
\end{table}

\begin{figure}[!t]
    \includegraphics[width=8cm]{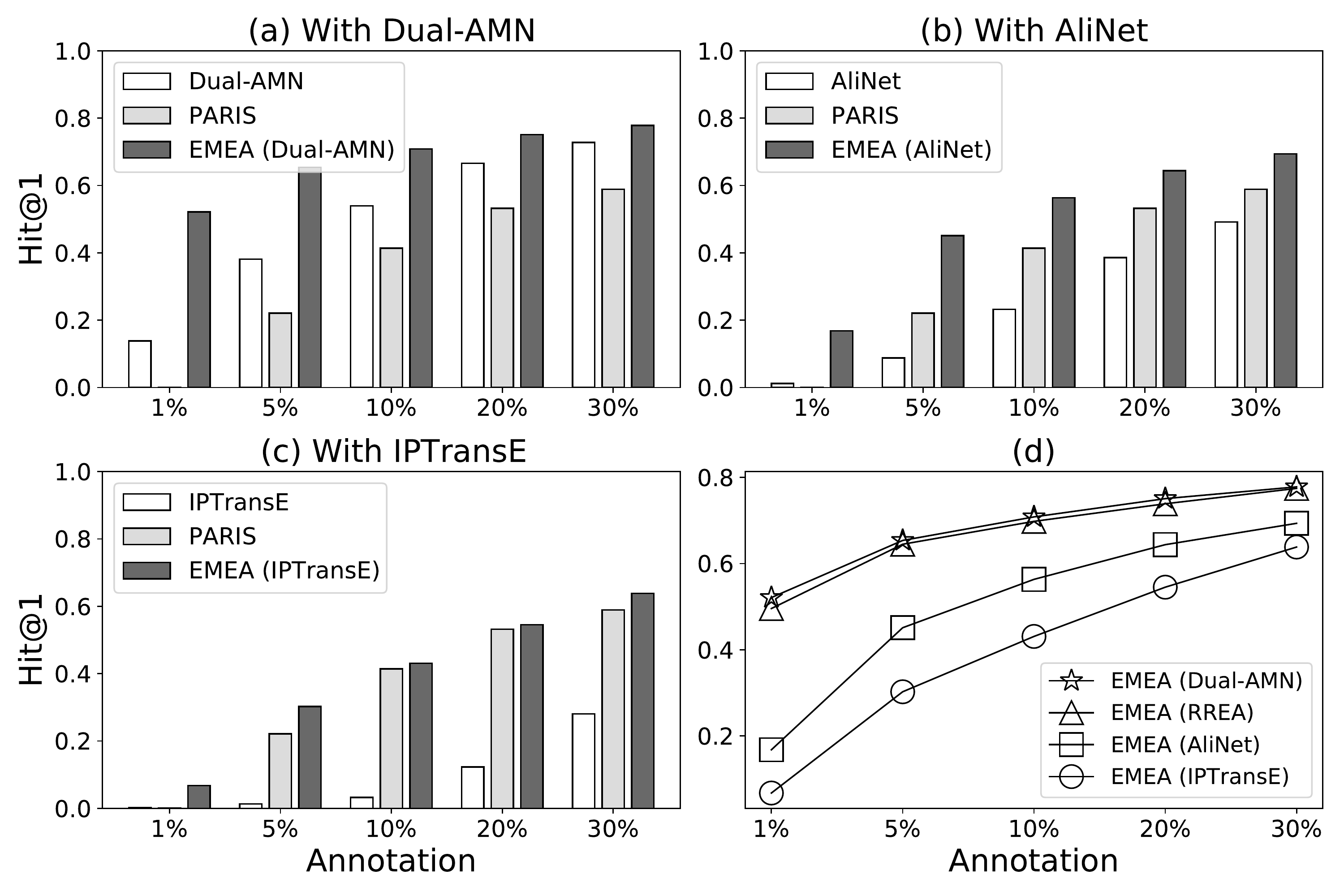}
    \caption{Generality of \textsf{EMEA} on neural EA models. Plots (a), (b) and (c) show \emea can boost different EA models consistently; Plot (d) shows better EA models lead to a better final performance.}
    \label{fig:generality_neural}
\end{figure}

\paragraph{Generality across Neural EA Models}
To explore the generality of \textsf{EMEA} across neural EA models, 
we apply \textsf{EMEA} to another three models: Dual-AMN~\cite{DBLP:conf/www/MaoWWL21}, AliNet~\cite{DBLP:conf/aaai/SunW0CDZQ20} and IPTransE~\cite{DBLP:conf/ijcai/ZhuXLS17} other than RREA. 
We find that:
(1)  \textsf{EMEA} can bring improvements to the three EA models as shown in Fig.\ref{fig:generality_neural} (a), (b) and (c).
Also, no matter whether the neural model is better or worse than PARIS, their combination using \textsf{EMEA} is more effective than using them separately.
(2) Fig.\ref{fig:generality_neural} (d) shows more effective neural models lead to more accurate final results.

\paragraph{Generality across Training Modes}
To verify the generality of \textsf{EMEA} across training modes, we also attempt to apply \textsf{EMEA} to improve semi-supervised RREA.
As reported in Table~\ref{tab:overall_perf_iterative}, we find that \textsf{EMEA} can also boost semi-supervised RREA consistently across different datasets and amounts of training data.
By comparing \emea in Table~\ref{tab:overall_perf} and Table~\ref{tab:overall_perf_iterative}, we find that semi-supervised RREA usually leads to better final EA effectiveness than supervised RREA after the boost of \emea.
Nevertheless, when the training data is extremely small, i.e. 1\%, semi-supervised RREA get worse final EA effectiveness than the supervised one. This might be caused by the low-quality seeds iteratively added into the training data during the semi-supervised training. We treat the exploration of this phenomenon as future work.

\section{Further Analysis}
Further analysis of \emea is done on dataset \textit{zh\_en}.

\paragraph{Impact of Compatibility on Training Process}

\begin{figure}
	\includegraphics[width=8cm]{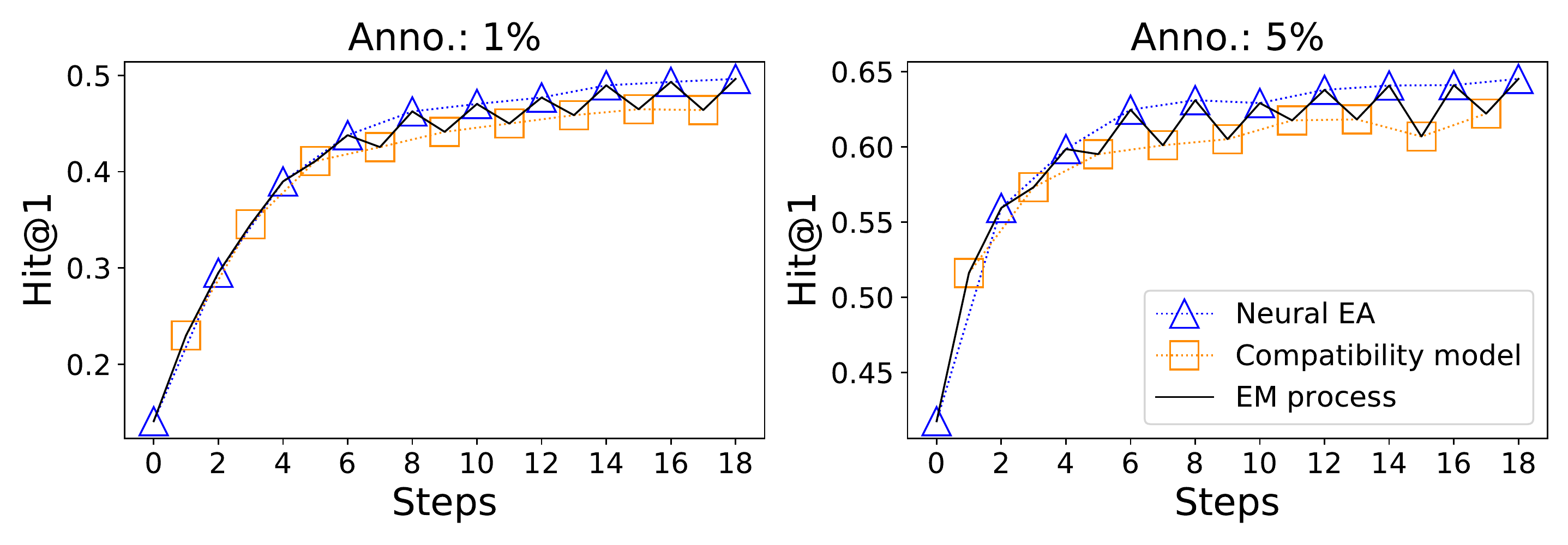}
	\caption{Convergence process of \textsf{EMEA} trained with 1\% and 5\%  annotations. Compatibility model can boost the neural EA model, no matter whether it can derive more accurate supervision signals or not.}
	\label{fig:overall_em_proc}
\end{figure}

In each iteration of the training procedure, the compatibility model derives more compatible predictions to assist in updating the neural EA model in the next iteration. To gain insights into their interaction, we examine the EA effectiveness of the two components in the training process. 
Fig.~\ref{fig:overall_em_proc} shows this processes running on datasets with 1\% and 5\% annotated data. 
Step 0 is the original neural EA model.
%
We make the following observations: 
(1) Overall, the effectiveness of the two components jointly increases across the training procedure. This phenomenon indicates that they indeed both benefit from each other.
(2) The compatibility model performs better than the neural model in the early stage while worse later. 
Thus, we can conclude that the compatibility model can always boost the neural EA model, regardless of whether it can derive more accurate supervision signals.

\paragraph{Effect of Rule Sets on \emea}

\begin{figure}
    \includegraphics[width=8cm]{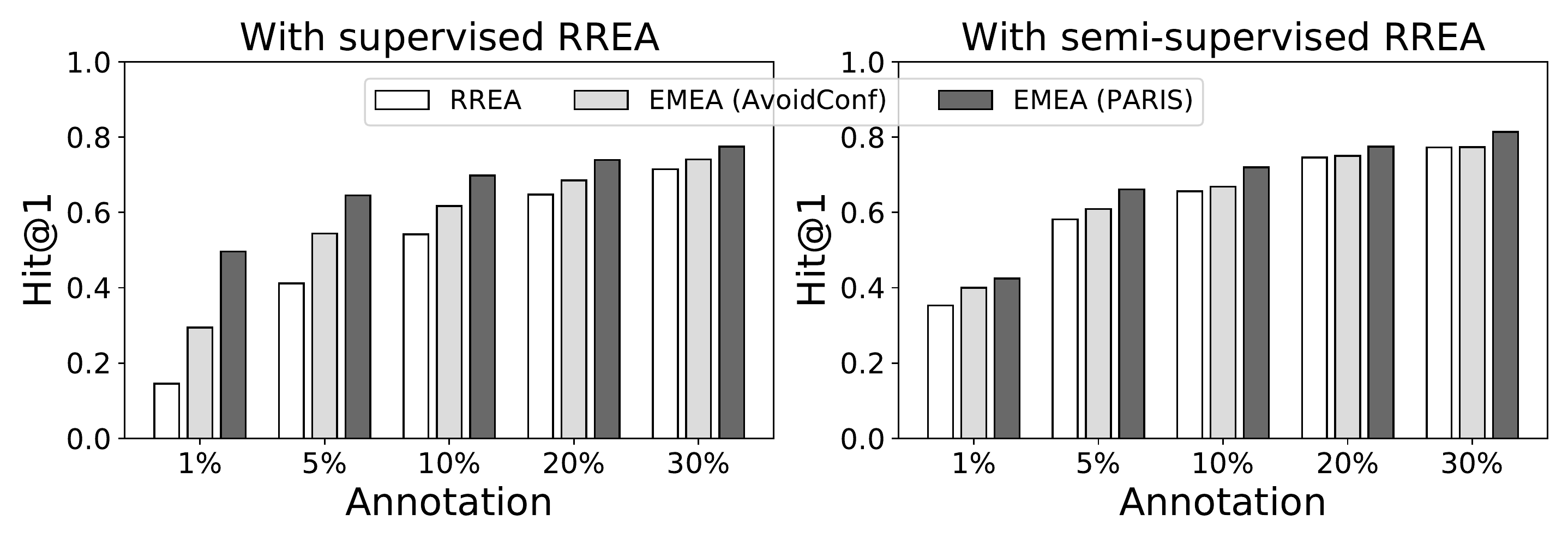}
    \caption{Effect of different rule sets on \textsf{EMEA}. PARIS rule is stronger than \textit{AvoidConf}.}
    \label{fig:effect_rules}
\end{figure}

To explore the effect of different rule sets on \emea, we deploy another rule set used for avoiding conflicts (denoted as \textit{AvoidConf} here) in \emea.
As shown in Fig.~\ref{fig:effect_rules}, \textit{AvoidConf} can improve supervised RREA consistently, as well as semi-supervised RREA when the training data is of a small amount. However, it can only bring very slight (or even no) improvement to the semi-supervised RREA when the training data is $>10\%$.
Also, its performance is always worse than PARIS rule.
Therefore, PARIS rule is stronger in indicating the compatibility of EA than \textit{AvoidConf}.
This finding is intuitively reasonable because the PARIS rule can be used to infer new mappings, while \textit{AvoidConf} can only indicate the potential errors.

\paragraph{Sensitivity to Hyper-parameters}\label{sec:hyperparameter}

\begin{figure}[t!]
    \includegraphics[width=8cm]{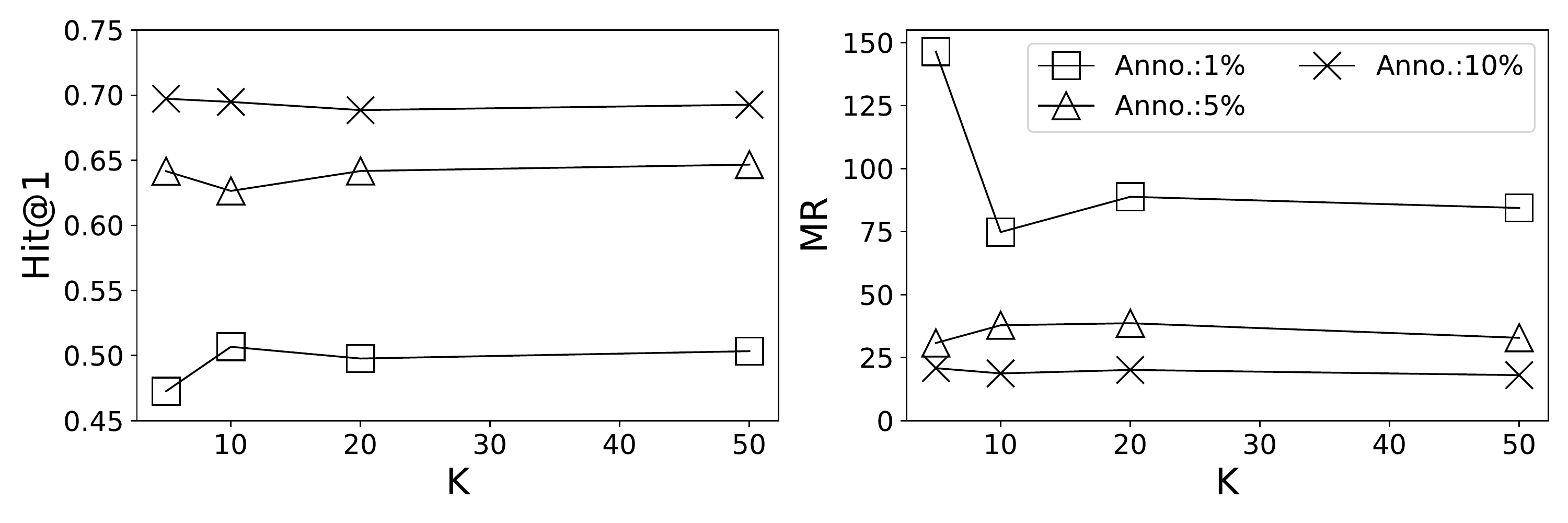}
    \caption{Sensitivity of \textsf{EMEA} to the parameter $K$ w.r.t. both shallow and deep metrics. The \emea is not sensitive when $K$ is not too small; The poorer EA model (trained with fewer data) is relatively more sensitive; A small $K$ out of the sensitive range is suggested for the trade-off between computation cost and EA effectiveness.}
    \label{fig:effect_of_topk}
\end{figure}

We also analyse the sensitivity of parameter $K$ (discussed in Sec. 4.5) in \emea under three annotation settings: 1\%, 5\% and 10\%.
Fig.~\ref{fig:effect_of_topk} shows the performance of \textsf{EMEA}, measured with Hit@1 and MR, with respect to different $K$ values. We find that: (1) the performance of \textsf{EMEA} fluctuates when $K<10$ but is relatively stable when $K>10$. (2) \textsf{EMEA} is more sensitive to $K$ in settings with fewer annotations. 
The reason is that a well-trained neural model can rank most true correspondences in the top positions and thus is less affected by the cut-off.
(3) Large $K$ values (e.g., 50) only make \textsf{EMEA} slightly more effective (note lower MR is better). Since larger $K$ values require more computation, a small value outside the sensitive range is suggested, like 10 in Fig.~\ref{fig:effect_of_topk}. 

\section{Conclusion}

Entity Alignment is a primary step of fusing different Knowledge Graphs.
Many neural EA models have been explored and achieved SOTA performance. 
Though, one nature of graph data -- the dependencies between entities -- is under-explored.
In this work, we raise attention to one neglected aspect of EA task -- different entities have compatible counterparts w.r.t. their underlying dependencies.
We argue that making self-consistent predictions should be one objective of training EA model other than fitting the labelled data.
Towards this goal, we devise one training framework named \emea, which can intervene in the update of EA model to improve its compatibility.
In \emea, we address three key problems: 
(1) measure compatibility with a graphical model which can aggregate local compatibilities;
(2) guide the training of EA models with the compatibility measure via variational inference;
(3) optimize the compatibility module with a variational EM framework. 
We empirically show that compatibility is very powerful in driving the training of neural EA models. The \emea complements the existing neural EA works.

\section*{Limitations}
Our \emea framework has a higher computation cost than the original neural EA model. It needs to compute the compatibility module and continue updating the neural EA model in the iterations. 
As a result, more time is taken to train the neural EA model.
In addition, we only measure compatibility in one direction (i.e. selecting one KG as the source KG and measuring compatibility on the target KG). Considering both directions might be able to further increase the EA performance. 

In future, we plan to explore dual compatibility modules. Also, we will study how to combine compatibility and self-training.

\section*{Acknowledgements}
This research is supported by the National Key Research and Development Program of China No. 2020AAA0109400, the Shenyang Science and Technology Plan Fund (No. 21-102-0-09), and the Australian Research Council (No. DE210100160 and DP200103650).

\bibliography{anthology,bibfile}
\bibliographystyle{acl_natbib}

\appendix


\section{Preliminary Study}\label{sec:preliminary}

\begin{figure}
	\centering
	\begin{minipage}{0.8\linewidth}
		\centering
		\includegraphics[width=5cm]{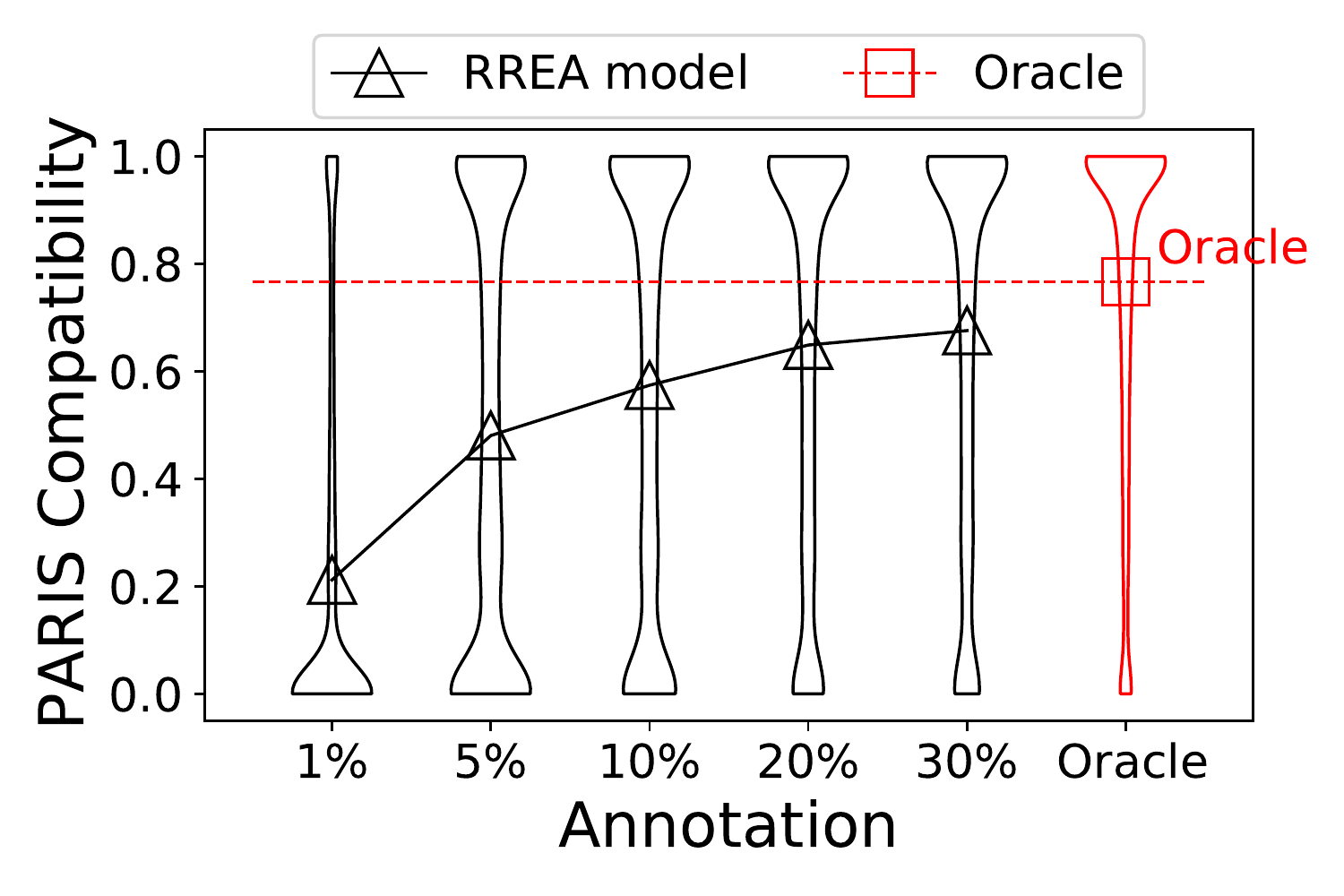}
		\text{(a) PARIS compatibility.}
	\end{minipage}
	\begin{minipage}{0.8\linewidth}
		\centering
		\includegraphics[width=5cm]{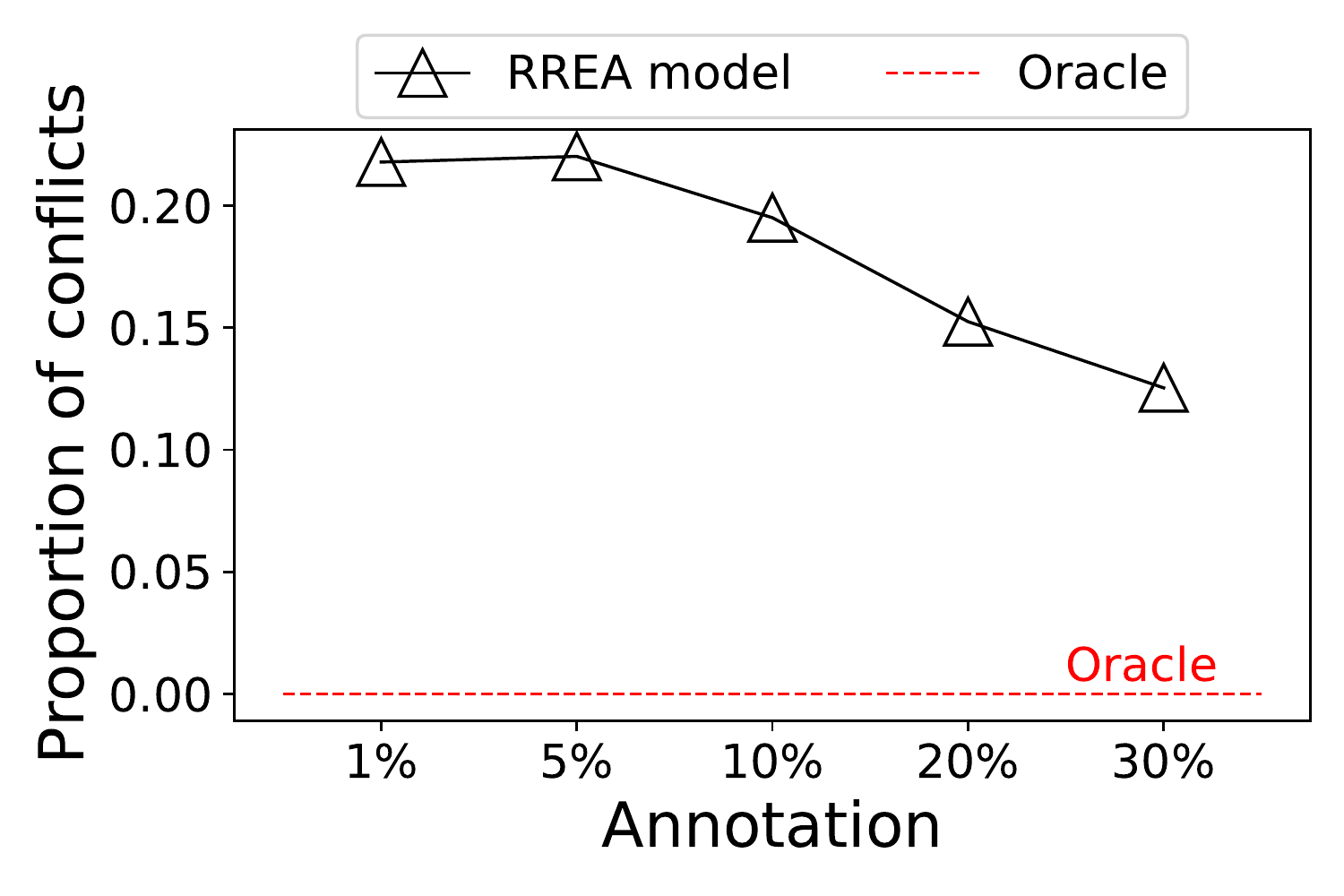}
		\text{(b) Conflicting prediction.}
	\end{minipage}
	\caption{PARIS compatibility and proportions of conflicting predictions of RREA trained with different proportions of annotated data. Both plots highlight the correlation between compatibility and performance in EA models.}
	\label{fig:compatibility_issue}
\end{figure}

Our preliminary study examines the compatibilities of neural EA models with different performances.
We choose RREA~\cite{DBLP:conf/cikm/MaoWXWL20} as the neural EA model and train it with different amounts of labelled data.
As for the compatibility, we measure it with PARIS rule and the number of conflicting predictions (see Sec.~\ref{sec:self-consistency}).
The PARIS compatibility is shown in Fig.~\ref{fig:compatibility_issue}, where the violins represent the distributions of compatibility scores while the triangles represent the average compatibilities. 
We can observe that RREA always makes numerous incompatible predictions, especially when trained with few annotated entities, while most oracle alignments are compatible. The curve strongly suggests that a better model makes more compatible predictions.
Similar findings can be observed in Fig.~\ref{fig:compatibility_issue} (b), where fewer conflicts mean better compatibility.
These observations motivate us to improve the neural EA methods by guiding them to better compatibility. 

\section{Derivation Processes}

\subsection{Simplification of $p_\Phi (y_e | y_{-e})$} \label{app:p_yi}

The derivation process of Eq.~\ref{eq:cond_p} is shown in Eq.~\ref{eq:cond_p_proc}.
$\mathrm{MB}^e$ is the Markov Blanket of $e$ and only contains entities cooccuring in any factor subset with $e$.
We only need to sample $y_{\mathrm{MB}^e}$ to compute $p_\Phi (y_e | y_{-e})$.

\begin{figure*}
    \begin{equation}
        \begin{aligned}
            p_\Phi (y_e | y_{-e}) & = \frac{p_\Phi(y_e, y_{-e})}{p_\Phi (y_{-e})} 
             = \frac{p_\Phi(y_e, y_{-e})}{\sum_{e' \in E'} p_\Phi(y_e=e', y_{-e})} \\
            & = \frac{\prod_{F | e \in F} \Phi(y_F) \times \prod_{F | e \notin F} \Phi(y_F)}{\sum_{e' \in E'} \prod_{F | e \in F} \Phi(y_F | y_e=e') \times \prod_{F | e \in F} \Phi(y_F)} \\
            & = \frac{\prod_{F | e \in F} \Phi(y_F) }{\sum_{e' \in E'} \prod_{F | e \in F} \Phi(y_F | y_e=e') }
            \doteq p_\Phi(y_e | y_{\mathrm{MB}^e})
        \end{aligned}
        \label{eq:cond_p_proc}
    \end{equation}
\end{figure*}

\subsection{Evidence Lower Bound}\label{app:elbo}

The relationship between KL divergence and ELBO shown in Eq.~\ref{eq:KL} and Eq.~\ref{eq:elbo} can be derived through Eq.~\ref{eq:elbo_proc}.

\begin{figure*}
    \begin{equation}
        \begin{split}
            \mathrm{KL}(q_\Theta(y_U) || p_\Phi(y_U|y_L)) & = \mathbb{E}_{q_\Theta(y_U )} \log \frac{q_\Theta(y_U)}{p_\Phi(y_U|y_L)}  \\
            & = \mathbb{E}_{q_\Theta(y_U)} \log q_\Theta(y_U)  - \mathbb{E}_{q_\Theta(y_U)} \log p_\Phi(y_U|y_L) \\
            & = \mathbb{E}_{q_\Theta(y_U)} \log q_\Theta(y_U) - \mathbb{E}_{q_\Theta(y_U)} \log p_\Phi(y_U, y_L) 
            + \mathbb{E}_{q_\Theta(y_U)} \log p_\Phi(y_L) \\
            & = -\left( \mathbb{E}_{q_\Theta(y_U)} \log p_\Phi(y_U, y_L) - \mathbb{E}_{q_\Theta(y_U)} \log q_\Theta(y_U) \right)  
            +  \log p_\Phi(y_L) 
        \end{split}
        \label{eq:elbo_proc}
    \end{equation}
\end{figure*}

\subsection{Derivation of $Q_\Theta$}\label{app:Q_proc}

The ELBO can be written as in Eq.~\ref{eq:q_proc}. In the last line, the first item is irrelevant to $\Theta$. Thus, we drop it and keep the second item as $Q_\Theta$.

\begin{figure*}
    \begin{equation}
        \begin{split}
        \mathrm{ELBO}&  = \mathbb{E}_{q_\Theta (y_U)} \log p_\Phi (y_U, y_L)
        - \mathbb{E}_{q_\Theta (y_U)} \log q_\Theta (y_U) \\
        & \approx \mathbb{E}_{q_\Theta (y_U)} \log  (p_\Phi (y_L) \prod_{u \in U} p_\Phi(y_u|y_{-u}, y_L) )  - \mathbb{E}_{q_\Theta(y_U)} \log \prod_{u \in U} q_\Theta (y_u) \\
        & = \mathbb{E}_{q_\Theta (y_U)} \log p_\Phi (y_L) + \mathbb{E}_{q_\Theta (y_U)} \sum_{i \in U} \log p_\Phi (y_u | y_{-u}, y_L) - \mathbb{E}_{q_\Theta (y_U)} \sum_{u \in U} \log q_\Theta (y_u) \\
        & = \log p_\Phi (y_L) +  \sum_{u \in U} \left( \mathbb{E}_{q_\Theta(y_u)} [\mathbb{E}_{q_\Theta (y_{-u})} \log p_\Phi (y_u | y_{-u}, y_L) ] - \mathbb{E}_{q_\Theta (y_u)} \log q_\Theta (y_u) \right)\\
        & = \log p_\Phi (y_L) +  \sum_{u \in U} \left( \mathbb{E}_{q_\Theta(y_u)} [\mathbb{E}_{q_\Theta (y_{-u})} \log p_\Phi (y_u | y_{-u}, y_L) - \log q_\Theta (y_u) ] \right)
        \end{split}
        \label{eq:q_proc}
        \end{equation}
\end{figure*}

\subsection{Derivation of $q^*$}\label{app:q_star_proc}

We write $Q_\Theta$ in the form of Eq.~\ref{eq:q_new_form}, and then differentiate ${Q_\Theta}$ regarding $q_\Theta (y_u)$ as in Eq.~\ref{eq:diff_Q}. By letting $\frac{d Q_\Theta}{d q_\Theta (y_u)}=0$, we can get Eq.~\ref{eq:q_star}. 
Note that the obtained $q^*$ in Eq.~\ref{eq:q_star} needs to be normalized into probabilities since it is proportional to (i.e.  $\propto$) the right size, which is not normalized. 

\begin{figure*}
    \begin{equation}
        Q_\Theta = \sum_{u \in U} \sum_{y_u} q_\Theta (y_u) \left( \mathbb{E}_{q_\Theta (y_{-u})} \log p_\Phi (y_u | y_{-u}, y_L) - \log q_\Theta (y_u) \right) 
        \label{eq:q_new_form}
    \end{equation}
    
    \begin{equation}
    \begin{aligned}
    \frac{d Q_\Theta}{d q_\Theta(y_u)} & = \left( \mathbb{E}_{q_\Theta (y_{-u})} \log p_\Phi (y_u | y_{-u}, y_L) - \log q_\Theta (y_u) \right)  + q_\Theta (y_u) (- \frac{1}{q_\Theta (y_u)}) \\
    &= \mathbb{E}_{q_\Theta (y_{-u})} \log p_\Phi (y_u | y_{-u}, y_L) - \log q_\Theta (y_u) - 1
    \end{aligned}
    \label{eq:diff_Q}
    \end{equation}


\end{figure*}




\subsection{Indicator Function of PARIS Rule}\label{app:paris}

With symbols used in this work, the probabilistic form of PARIS rule, i.e.  our indicator function, can be written as Eq.~\ref{eq:paris_prob}, where $r(e,n)$ denote any triple with $e$ as head entity, $r'(y_e, n')$ denote any triple with $y_e$ as head entity. In addition, $\Pr (r' \subseteq r )$ represents the likelihood that $r'$ is a subrelation of $r$ ($\Pr (r \subseteq r' )$ is analogous), while $fun^{-1}(r)$ denotes the reverse functionality of relation $r$. See \cite{DBLP:journals/pvldb/SuchanekAS11}  for more details about them.

\begin{equation}
    \begin{aligned}
        &g(y_{F_{e}}) = \Pr (e \equiv y_e) =  1 - \prod_{\substack{r(e,n), r'(y_e,n')}} \\
        & \left(1 - \Pr (r' \subseteq r ) \times fun^{-1}(r) \times q_\Theta (y_{n} = n') \right) 
        \times  \\
        & \left(1 - \Pr (r \subseteq r' ) \times fun^{-1}(r') \times q_\Theta (y_{n} = n') \right)
    \end{aligned}
    \label{eq:paris_prob}
\end{equation}


\section{Experiments}\label{app:exp}

\subsection{Performance of Neural EA Models}\label{app:neural_ea_models}

Table~\ref{tab:baselines} summarizes the performance of SOTA neural EA models.
The results of IPTransE, GCN\-Align, MUGNN, RSN, AliNet are reported in \cite{DBLP:conf/aaai/SunW0CDZQ20} while others are reported in their original papers.
The results of RREA trained with 30\% of training data in Table~\ref{tab:overall_perf} are reproduced by us and slightly different from the results reported in Table~\ref{tab:baselines} because of different random settings. Similar situation can be found in Table~\ref{tab:overall_perf_iterative}.

\begin{table*}
    \centering
    \caption{Performance of neural EA baselines. Percentage of labelled data: 30\%; \textit{sup}: supervised; \textit{semi}: semi-supervised.}
    \scalebox{0.8}[0.8]{
    \begin{tabular}{|c|c|c|c|c|c|c|}
    \hline
    \multirow{2}{*}{Method} & \multicolumn{2}{c|}{zh\_en} & \multicolumn{2}{c|}{ja\_en} & \multicolumn{2}{c|}{fr\_en} \\
     & Hit@1 & MRR & Hit@1 & MRR & Hit@1 & MRR \\
    \hline
    IPTransE~\cite{DBLP:conf/ijcai/ZhuXLS17} & 0.406 & 0.516 & 0.367 & 0.474 & 0.333 & 0.451 \\
    GCN-Align~\cite{DBLP:conf/emnlp/WangLLZ18} & 0.413 & 0.549 & 0.399 & 0.546 & 0.373 & 0.532 \\
    MuGNN~\cite{DBLP:conf/acl/CaoLLLLC19} & 0.494 & 0.611 & 0.501 & 0.621 & 0.495 & 0.621 \\
    RSN~\cite{DBLP:conf/icml/GuoSH19} & 0.508 & 0.591 & 0.507 & 0.590 & 0.516 & 0.605 \\
    AliNet~\cite{DBLP:conf/aaai/SunW0CDZQ20} & 0.539 & 0.628 & 0.549 & 0.645 & 0.552 & 0.657 \\
    MRAEA (sup)~\cite{DBLP:conf/wsdm/MaoWXLW20} & 0.638 & 0.736 & 0.646 & 0.735 & 0.666 & 0.765 \\
    PSR (sup)~\cite{DBLP:conf/cikm/MaoWWL21} &  0.702 & 0.781 & 0.698 & 0.782 & 0.731 & 0.807 \\
    RREA (sup)~\cite{DBLP:conf/cikm/MaoWXWL20} &  0.715 & 0.794 & 0.713 & 0.793 & 0.739 & 0.816 \\
    Dual-AMN (sup)~\cite{DBLP:conf/www/MaoWWL21} &  0.731 & 0.799 & 0.726 & 0.799 & 0.756 & 0.827 \\
    \hline
    BootEA (semi)~\cite{DBLP:conf/ijcai/SunHZQ18} & 0.629 & 0.703 & 0.622 & 0.701 & 0.653 & 0.731 \\
    MRAEA (semi)~\cite{DBLP:conf/wsdm/MaoWXLW20}& 0.757 & 0.827 & 0.758 & 0.826 & 0.781 & 0.849 \\
    PSR (semi)~\cite{DBLP:conf/cikm/MaoWWL21} &  0.802 & 0.851 & 0.803 & 0.852 & 0.828 & 0.874 \\
    RREA (semi)~\cite{DBLP:conf/cikm/MaoWXWL20} & 0.801 & 0.857 & 0.802 & 0.858 & 0.827 & 0.881 \\
    Dual-AMN (semi)~\cite{DBLP:conf/www/MaoWWL21} &  0.808 & 0.857 & 0.801 & 0.855 & 0.840 & 0.888 \\
    \hline
\end{tabular}
    }
    \label{tab:baselines}
\end{table*}

\begin{table}[t]
    \caption{Time consumption (seconds) of running EMEA on zh\_en.}
    \scalebox{0.77}[0.77]{
    \begin{tabular}{|c|c|c|c|}
    \hline
    Anno. & Initialization & Neural Module & Joint Distr. Module \\
    \hline
    1 \%& 254.1 & 108.1$\pm$1.2 & 88.5$\pm$1.1 \\
    \hline
    5 \%& 249.7 & 108.0$\pm$2.5 & 98.2$\pm$15.5 \\
    \hline
    10 \%& 249.2 & 108.6$\pm$0.9 & 86.0$\pm$0.3 \\
    \hline
    20 \%& 252.8 & 109.6$\pm$1.6 & 86.7$\pm$0.6 \\
    \hline
    30 \%& 252.4 & 111.1$\pm$2.0 & 90.2$\pm$2.7 \\
    \hline
\end{tabular}
    }
    \label{tab:efficiency15k}
\end{table}

\begin{table}[t]
    \caption{Time consumption (minutes) of running EMEA on dbp\_wd.}
    \scalebox{0.77}[0.77]{
    \begin{tabular}{|c|c|c|c|}
    \hline
    Anno. & Initialization & Neural Module & Joint Distr. Module \\
    \hline
    1 \%& 169.1 & 36.8$\pm$0.6 & 19.1$\pm$1.4 \\
    \hline
    5 \%& 172.2 & 33.0$\pm$6.6 & 15.9$\pm$3.9 \\
    \hline
    10 \%& 146.3 & 43.5$\pm$3.2 & 22.8$\pm$2.4 \\
    \hline
    20 \%& 190.0 & 45.3$\pm$3.3 & 29.2$\pm$2.0  \\
    \hline
    30 \%& 183.9 & 49.0$\pm$4.4 & 30.2$\pm$6.9  \\
    \hline
\end{tabular}

    }
    \label{tab:efficiency100k}
\end{table}

\subsection{Details for Reproducibility}\label{app:repro}
\paragraph{Hyper-parameters}
We search the number of candidate counterparts $K$ from [5,10,20,50], and set it as 10 for the trade-off of effectiveness and efficiency as discussed in Sec.~\ref{sec:hyperparameter}.
The number of nearest neighbours $N$ in the \textit{AvoidConf} rule is searched from [5,10,15,20] and set as 5 for similar consideration as $K$;

\paragraph{Implementation of Baselines and Neural EA Models}
The source codes of PARIS~\footnote{\url{https://github.com/dig-team/PARIS}} and PRASE~\footnote{\url{https://github.com/qizhyuan/PRASE-Python}} are used to produce their results.
As for the neural EA models, RREA~\footnote{\url{https://github.com/MaoXinn/RREA}} and Dual-AMN~\footnote{\url{https://github.com/MaoXinn/Dual-AMN}} are implemented based on their source codes, while AliNet and IPTransE are implemented with OpenEA.~\footnote{\url{https://github.com/nju-websoft/OpenEA}}.
We use the default settings of their hyper-parameters in these source codes.

\paragraph{Configuration of Running Device}
We run the experiments on one GPU server, which is configured with an Intel(R) Xeon(R) Gold 6128 3.40GHz CPU, 128GB memory, 3 NVIDIA GeForce GTX 2080Ti GPU and Ubuntu 20.04 OS.

\subsection{Running Time}

We report our running time on datasets {zh\_en} (15K)  and {dbp\_wd} (100K) in Table~\ref{tab:efficiency15k} and \ref{tab:efficiency100k} as reference.
The experiments on other datasets take comparable running time as them w.r.t. corresponding dataset size.
Note that these experiments are run on a shared server and cannot be used to measure the precise running efficiency.
In each table, we count the time consumption of initializing EA model, updating EA model and computing the self-consistency module in each EM iteration.
The experiments on dbp\_wd (100K) are slow because the source code of RREA can only run on CPU while raising Out-of-Memory exception on GPU.


\end{document}